\documentclass[letterpaper, 10 pt, conference]{ieeeconf}  

\IEEEoverridecommandlockouts                              
\usepackage{graphicx}
\usepackage{amsmath}
\usepackage{url}
\usepackage{cite}

\DeclareMathOperator*{\argmax}{argmax}

\renewcommand{\baselinestretch}{1.0}

\usepackage{algorithm}
\usepackage{algpseudocode}
\algrenewcommand{\algorithmicreturn}{\State \textbf{return}}

\title{\LARGE \bf 
Online Spatial Concept and Lexical Acquisition \\with Simultaneous Localization and Mapping
}

\author{Akira Taniguchi$^{1}$, Yoshinobu Hagiwara$^{1}$, Tadahiro Taniguchi$^{1}$ and Tetsunari Inamura$^{2}$
\thanks{*This work was partially supported by JST, CREST. }
\thanks{$^{1}$Akira Taniguchi, Yoshinobu Hagiwara and Tadahiro Taniguchi are with Ritsumeikan University, 1-1-1 Noji-Higashi, Kusatsu, Shiga 525-8577, Japan
        {\tt\small \{a.taniguchi, yhagiwara, taniguchi\} @em.ci.ritsumei.ac.jp}}%
\thanks{$^{2}$Tetsunari Inamura is with National Institute of Informatics / SOKENDAI (The Graduate University for Advanced Studies),
        2-1-2 Hitotsubashi, Chiyoda-ku, Tokyo 101-8430, Japan
        {\tt\small inamura@nii.ac.jp}}%
}

\begin{document}

\maketitle
\thispagestyle{empty}
\pagestyle{empty}

\begin{abstract}
In this paper, we propose an online learning algorithm based on a Rao-Blackwellized particle filter for spatial concept acquisition and mapping.
We have proposed a nonparametric Bayesian spatial concept acquisition model (SpCoA).
We propose a novel method (SpCoSLAM) integrating SpCoA and FastSLAM in the theoretical framework of the Bayesian generative model.
The proposed method can simultaneously learn place categories and lexicons while incrementally generating an environmental map.
Furthermore, the proposed method has scene image features and a language model added to SpCoA.
In the experiments, we tested online learning of spatial concepts and environmental maps in a novel environment of which the robot did not have a map.
Then, we evaluated the results of online learning of spatial concepts and lexical acquisition.
The experimental results demonstrated that the robot was able to more accurately learn the relationships between words and the place in the environmental map incrementally by using the proposed method.
\end{abstract}

\section{INTRODUCTION}
\label{sec:introduction}
Robots coexisting with humans and operating in various environments are required to adaptively learn and use the spatial concepts and vocabulary related to different places.
However, spatial concepts are such that their target domain may be unclear compared with object concepts and may differ according to the user and environment.
Therefore, it is difficult to manually design spatial concepts in advance, and it is desirable for robots to autonomously learn spatial concepts based on their own experiences.

The related research fields of semantic mapping and place categorization~\cite{kostavelis2015semantic,landsiedel2017review} have attracted considerable interest in recent years.
However, most of these studies have consisted of separate independent methods of semantics of places and mapping using simultaneous localization and mapping (SLAM)~\cite{thrun2005probabilistic}.
In addition, the semantics of places, place categories, and names of places could only be learned from pre-set values.
In this paper, we propose a novel unsupervised Bayesian generative model and an online learning algorithm that can perform simultaneous learning of the spatial concepts and an environmental map from multimodal information.
The proposed method can automatically and sequentially perform place categorization and learn unknown words without prior knowledge.

\begin{figure}[tb]
  \begin{center}
    \includegraphics[width=1.00\hsize]{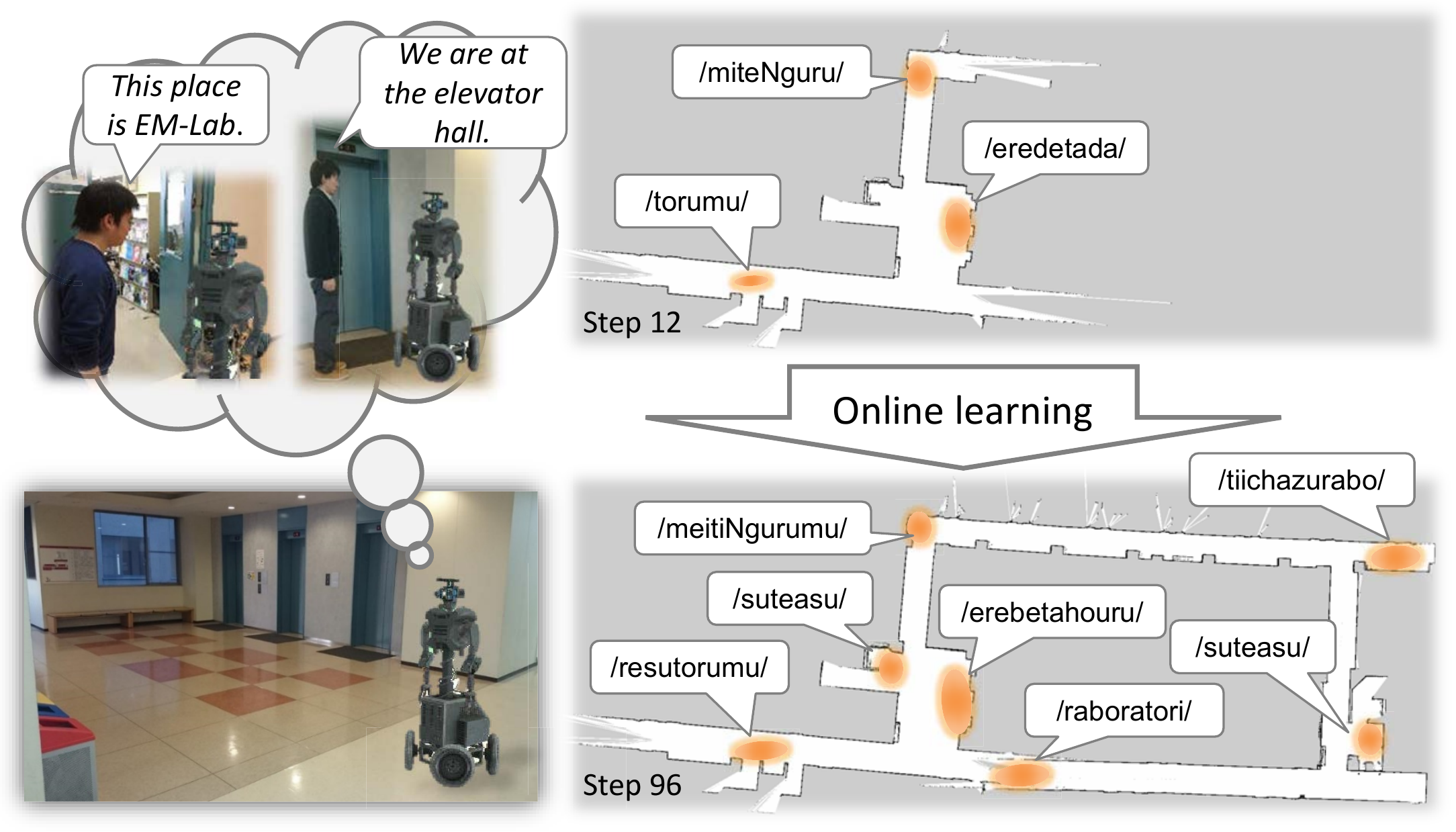}
    \caption{Overview of online learning of spatial concepts and an environmental map; We aim to develop a method that enables mobile robots to learn spatial concepts, a lexicon and an environmental map sequentially from interaction with an environment and human, even in an unknown environment without prior knowledge.}
    \label{fig:overview}
  \end{center}
\end{figure}

Taniguchi et al.~\cite{ataniguchi_IFAC2016} proposed a method that integrated ambiguous speech-recognition results with the self-localization method for learning spatial concepts.
In addition, Taniguchi et al.~\cite{taniguchi_spcoa} proposed the nonparametric Bayesian spatial concept acquisition method~(SpCoA) based on an unsupervised word-segmentation method known as latticelm~\cite{neubig2012bayesian}.
On the other hand, Ishibushi et al.~\cite{ishibushi2015statistical} proposed a self-localization method that exploits image features using a convolutional neural network (CNN)~\cite{krizhevsky2012imagenet}.
These methods~\cite{ataniguchi_IFAC2016,taniguchi_spcoa,ishibushi2015statistical} cannot cope with changes in the names of places and the environment because these methods use batch learning algorithms.
In addition, these methods cannot learn spatial concepts from unknown environments without a map, i.e., the robot needs to have a map generated by SLAM beforehand.
Therefore, in this paper, we develop an online algorithm that can sequentially learn a map, spatial concepts integrating positions, speech signals, and scene images. 

FastSLAM~\cite{gridbasedfastslam2005,gridbasedfastslam2007} has realized an on-line algorithm for efficient self-localization and mapping using a Rao-Blackwellized particle filter (RBPF)~\cite{doucet2000rao}.
In this paper, we introduce a grid-based FastSLAM algorithm in the generative model for spatial concept acquisition.
The graphical model of SpCoA has integrated spatial lexical acquisition into Monte Carlo localization (MCL), a particle-filter-based self-localization method.
SpCoA can be extended naturally to SLAM.
Therefore, we assume that the robot can learn vocabulary related to places and a map sequentially.

One of the important problems of our research is unsupervised lexical acquisition.
There are research efforts on incremental spatial language acquisition through robot-to-robot interaction~\cite{Spranger2015,heathlingodroids2016}.
However, these studies~\cite{Spranger2015,heathlingodroids2016} did not consider lexical acquisition through human-to-robot speech interactions (HRSI).
For online unsupervised lexical acquisition by HRSI, it is necessary to deal with the problems of phoneme recognition errors and word segmentation of uttered sentences containing errors.
SpCoA reduced phoneme recognition errors of word segmentation by using the weighted finite-state transducer (WFST)-based unsupervised word segmentation method latticelm~\cite{neubig2012bayesian}.
Araki et al.~\cite{araki2012online} performed a pseudo-online algorithm using the nested Pitman--Yor language model (NPYLM)~\cite{mochihashi2009bayesian_short}.
However, these studies~\cite{taniguchi_spcoa,araki2012online} have reported that word segmentation of speech recognition results including errors causes over-segmentation~\cite{goldwater2009bayesian}.
In this paper, we will improve the accuracy of speech recognition by updating the language models sequentially.

We assume that the robot has not acquired any vocabulary in advance, and can recognize only phonemes or syllables.
We represent the spatial area of the environment in terms of a {\it position distribution}.
Furthermore, we define a {\it spatial concept} as a place category that includes place names, scene image features, and the position distributions corresponding to those names.

The goal of this study is to develop a robot that learns spatial concepts incrementally from multimodal information obtained while moving in the environment.
The main contributions of this paper are as follows.
\begin{itemize}
 \item       We propose an online algorithm based on RBPF for spatial concept acquisition.
 The proposed method integrates SpCoA and FastSLAM in the theoretical framework of the Bayesian generative model.
 \item       We demonstrated that a robot without a pre-existing lexicon or map can learn spatial concepts and an environmental map incrementally.
\end{itemize}

\section{Online Spatial Concept Acquisition}
\label{sec:SpCoSLAM}
\subsection{Overview}
\label{sec:SpCoSLAM:overview}
An overview of the proposed method is shown in Fig.~\ref{fig:overview}.
The proposed method is an online spatial concept acquisition and simultaneous localization and mapping (SpCoSLAM).
The proposed method can learn sequential spatial concepts for unknown environments and unsearched regions without maps.
In addition, it can mutually complement the uncertainty of information by using multimodal information. 
A pseudo-code for the online learning is given in Algorithm~\ref{alg:SpCoSLAM}.
The procedure of SpCoSLAM for each step is described as follows.
2) -- 6) are performed for each particle.
\begin{enumerate}
\item
A robot gets WFST speech recognition results of the user's speech signals using a language model of the previous step. (line 3 in Algorithm~\ref{alg:SpCoSLAM})
\item
The robot gets the observation likelihood by performing a sample motion model and a measurement model of FastSLAM. (line 5-10)
\item
The robot performs unsupervised word segmentation latticelm~\cite{neubig2012bayesian} using WFST speech recognition results. (line 11)
\item
The robot gets latent variables of spatial concepts by sampling.
The details of this process are described in Section~\ref{sec:SpCoSLAM:sampling}. (line 12)
\item
The robot gets the marginal likelihood of observation data as the importance weight. (line 13-15)
\item
The robot updates an environmental map. (line 16)
\item
The robot estimates the set of parameters of spatial concepts from data and sampled values. (line 17)
\item
The robot updates a language model of the maximum weight for next step. (line 20-21)
\item
The robot performs resampling of particles according to weights. (line 22-25)
\end{enumerate}

\begin{figure}[tb]
  \begin{center}
    \includegraphics[width=160pt]{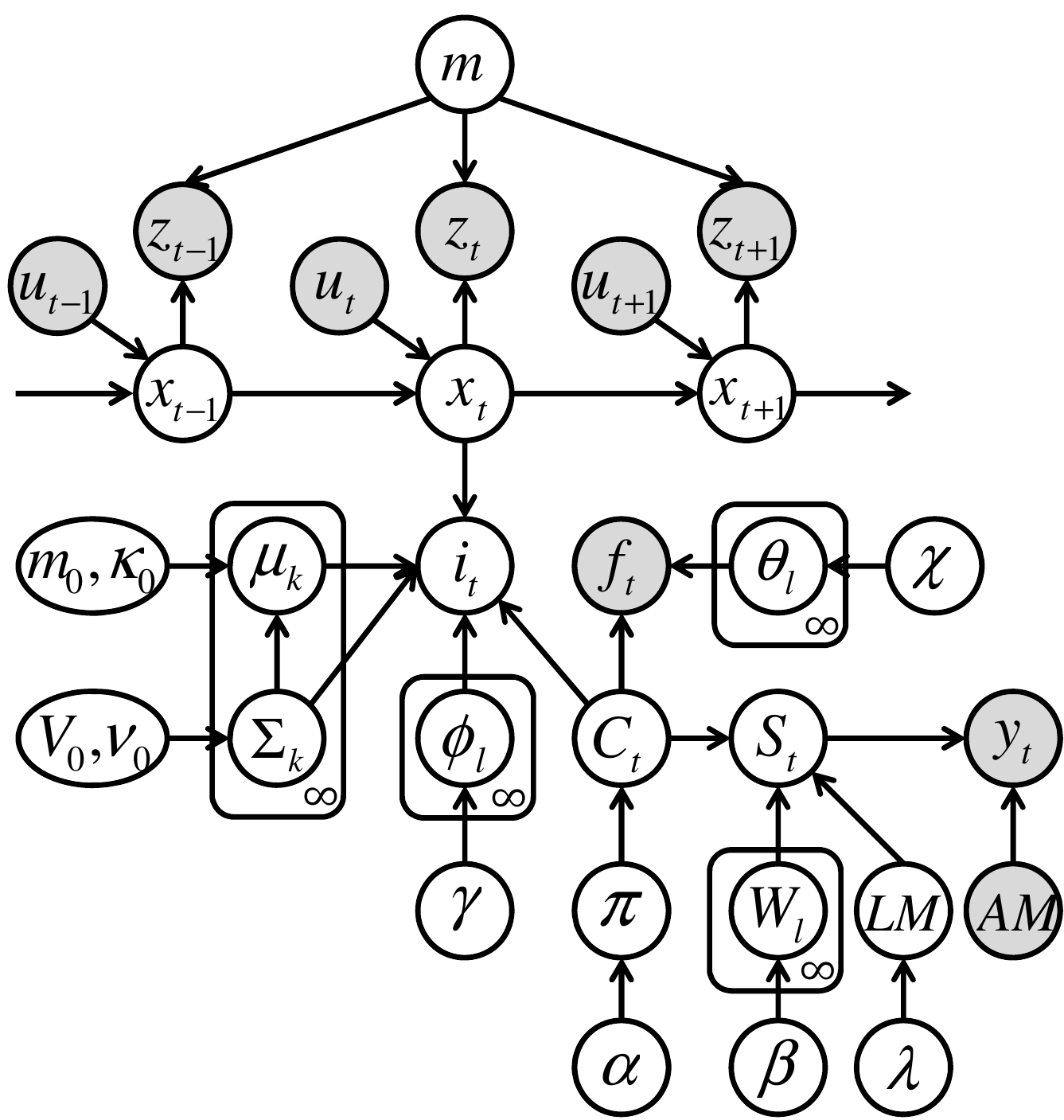}
    \caption{Graphical model representation of SpCoSLAM; It expresses multimodal place categorization, lexical acquisition and SLAM as one Bayesian generative model. Gray nodes indicate observation variables.} 
    \label{fig:gmodel}
  \end{center}
\end{figure}

\begin{table}[tb]
\begin{center}
\caption{Each element of the graphical model of SpCoSLAM}
\begin{tabular}{c|c} \hline
$x_t$ & Self-position of a robot \\ \hline
$z_t$ & Sensor data (depth data) \\ \hline
$u_t$ & Control data \\ \hline
$f_{t}$ & Image feature \\ \hline
$y_{t}$ & Speech signal \\ \hline
\raisebox{1ex}{\shortstack{$S_t$}} & {\shortstack{\raisebox{-8pt}{Sequence of words} \\ (word segmentation result)}}  \\ \hline
$C_{t}$ & Index of spatial concepts \\ \hline
$i_{t}$ & Index of position distributions \\ \hline
$m$ & Environmental map \\ \hline
\raisebox{1ex}{\shortstack{$\pi $}} & {\shortstack{\raisebox{-8pt}{Multinomial distribution} \\of index $C_t$ of spatial concepts}} \\ \hline
\raisebox{1ex}{\shortstack{$\phi_{l}$}} & {\shortstack{\raisebox{-8pt}{Multinomial distribution} \\of index $i_{t}$ of position distribution}} \\ \hline
\raisebox{1ex}{\shortstack{$\mu_{k}$, $\Sigma_{k}$}} & {\shortstack{\raisebox{-8pt}{Position distribution}\\
(mean vector, covariance matrix)}} \\ \hline
\raisebox{1ex}{\shortstack{$\theta_{l}$}} & {\shortstack{\raisebox{-8pt}{Multinomial distribution} \\of image feature}} \\ \hline
\raisebox{1ex}{\shortstack{$W_{l}$}} & {\shortstack{\raisebox{-8pt}{Multinomial distribution} \\of the names of places}} \\ \hline
$LM$ & Language model (word dictionary) \\ \hline 
$AM$ & Acoustic model for speech recognition \\ \hline
\raisebox{-1ex}{\shortstack{$\alpha $,$\beta$,$\gamma $,$\chi $,$\lambda$ \\$m_{0}$,$\kappa _{0}$,$V_{0}$,$\nu _{0}$}}  & {\shortstack{\raisebox{1.0pt}{Hyperparameters of prior distributions}}}  \\ \hline
\end{tabular}
\label{youso2}
\end{center}
\end{table}

\subsection{Definition of generative model and graphical model}
Figure~\ref{fig:gmodel} shows the graphical model of SpCoSLAM and Table~\ref{youso2} lists each variable of the graphical model.
We describe the formulation of the generation process represented by the graphical model as follows:
\begin{eqnarray}
\pi &\sim& {\rm DP}(\alpha ) \label{eq:seisei1} \\
C_{t} &\sim& {\rm Mult}(\pi) \label{eq:seisei2} \\
\phi_{l} &\sim& {\rm DP}(\gamma ) \label{eq:seisei5} \\
W_{l} &\sim& {\rm Dir}(\beta) \label{eq:seisei3} \\
LM &\sim&  p(LM \mid \lambda) \label{eq2:seiseilm} \\
S_{t} &\sim& p(S_{t} \mid {\bf{W}}, C_{t},LM) \label{eq2:seisei4b} \\
y_{t} &\sim& p(y_{t} \mid S_{t},AM) \label{eq2:seiseiYt} \\
\theta_{l} &\sim& {\rm Dir}(\chi) \label{eq3:seisei3} \\
f_{t} &\sim& {\rm Mult}(\theta_{C_{t}}) \label{eq3:seisei2} \\
\Sigma_{k} &\sim& {\rm IW}( \Sigma \mid V_{0} ,\nu _{0} )  \label{eq:seisei7} \\
\mu_{k} &\sim& {\rm N}( \mu  \mid m_{0}, ( \Sigma_{k} / \kappa _{0} )) \label{eq:seisei8} \\
x_{t} &\sim& p(x_{t} \mid x_{t-1},u_{t}) \label{eq:seisei9} \\
z_{t} &\sim& p(z_{t} \mid x_{t},m) \label{eq:seisei10} \\
i_{t} &\sim& p(i_{t} \mid x_{t}, \mbox{\boldmath{$\mu $}}, \mbox{\boldmath{$\Sigma$}}, \mbox{\boldmath{$\phi$}}, C_{t}) \label{eq:seisei6} 
\label{eq:seisei}
\end{eqnarray}
where ${\rm DP()}$ represents Dirichlet process, ${\rm Mult()}$ is multinomial distribution, ${\rm Dir()}$ is Dirichlet distribution, ${\rm IW()}$ is inverse–Wishart distribution, and ${\rm N()}$ is Gaussian distribution.

Equation~(\ref{eq2:seisei4b}) approximates by using unigram rescaling~\cite{gildea1999topic}, as shown in (\ref{eq2:st_UR}).
$\overset{\mathrm{UR}}{\approx }$ represents the approximation by unigram rescaling.
\begin{eqnarray}
\lefteqn{
p(S_{t} \mid {\bf{W}}, C_{t},LM) 
}\quad \nonumber \\
&\overset{\mathrm{UR}}{\approx }& p(S_{t} \mid LM) \prod_{B_{t}} \frac{{\rm Mult}(S_{t,b} \mid W_{C_{t}})}{\sum_{c'}{\rm Mult}(S_{t,b} \mid W_{c'})}
\label{eq2:st_UR}
\end{eqnarray}
where $B_{t}$ denotes the number of words in the sentence.

Then, the probability distribution for (\ref{eq:seisei6}) can be defined as follows:
\begin{eqnarray}
\lefteqn{
p(i_{t} \mid x_{t},\mbox{\boldmath{$\mu $}},\mbox{\boldmath{$\Sigma$}} ,\mbox{\boldmath{$\phi$}}, C_{t})
}\quad \nonumber \\
&=& \cfrac{{\rm N}(x_{t} \mid \mu _{i_{t}}, \Sigma_{i_{t}}){\rm Mult}(i_{t} \mid \phi_{C_{t}})}{\sum_{i_{t}=j} {\rm N}(x_{t} \mid \mu _{j}, \Sigma_{j}){\rm Mult}(j \mid \phi_{C_{t}})}~.
\label{eq:it}
\end{eqnarray}

\subsection{Formulation of the speech recognition and the unsupervised word segmentation}
\label{sec:recognition_segmentation}
The 1-best speech recognition and the WFST speech recognition are represented as follows:
\begin{eqnarray}
S^{{\rm (1\mathchar`-best)}}_{t} &=& \argmax_{S_{t}} {\rm SR}(S_{t} \mid y_{t},AM,LM) \\
{\cal L}_{t} &\approx&  {\rm SR}({\cal L}_{t} \mid y_{t},AM,LM) 
\end{eqnarray}
where ${\cal L}_{t}$ denotes the speech recognition result of WFST format, which is a word graph representing the speech recognition results.
The unsupervised word segmentation of WFST by latticelm~\cite{neubig2012bayesian} is represented as follows:
\begin{eqnarray}
S_{T_{o}} &\sim& latticelm(S_{T_{o}} \mid {\cal L}_{T_{o}} , \lambda).
\end{eqnarray}

\begin{algorithm}[tb]  
\caption{Online learning algorithm of SpCoSLAM} 
\label{alg:SpCoSLAM}  
\small
\begin{algorithmic}[1]
\Procedure{${\rm SpCoSLAM}$}{$X_{t-1},u_{t},z_{t},f_{1:t},y_{1:t}$}
\State $\bar{X}_{t} = X_{t} =  \emptyset$
\State ${\cal L}_{1:t} = {\rm SR}({\cal L}_{1:t} \mid y_{1:t},AM,LM_{t-1})$
\For{$r=1$ to $R$}
\State $\acute{x}_{t}^{[r]} = {\rm\bf sample\_motion\_model}(u_{t},x_{t-1}^{[r]})$
\State ${x}_{t}^{[r]} = {\rm\bf scan\_matching}(z_{t}, \acute{x}_{t}^{[r]}, m_{t-1}^{[r]})$ 
\For{$j=1$ to $J$}
\State ${x}_{j} = {\rm\bf sample\_motion\_model}(u_{t},x_{t-1}^{[r]})$
\EndFor
\State $\omega_{z}^{[r]} = \sum_{j=1}^{J} {\rm\bf measurement\_model}(z_{t}, {x}_{j}, m_{t-1}^{[r]})$ 
\State $S_{1:t}^{[r]} \sim latticelm(S_{1:t} \mid {\cal L}_{1:t},\lambda)$
\State $i_{t}^{[r]},C_{t}^{[r]} \sim p(i_{t},C_{t} \mid x_{0:t}^{[r]}, i_{1:t-1}^{[r]},C_{1:t-1}^{[r]},S_{1:t}^{[r]},f_{1:t},\mathbf{h})$
\State $\omega_{f}^{[r]} = p(f_{t} \mid  C_{1:t-1}^{[r]}, f_{1:t-1}, \alpha,\chi)$
\State $\omega_{s}^{[r]} = {p(S_{t}^{[r]}\,|\,S_{1:t-1}^{[r]}, C_{1:t-1}^{[r]}, \alpha,\beta)}/{p(S_{t}^{[r]} \mid S_{1:t-1}^{[r]}, \beta)} $
\State $\omega_{t}^{[r]} = \omega_{z}^{[r]} \cdot \omega_{f}^{[r]} \cdot \omega_{s}^{[r]}$
\State $m_{t}^{[r]} = {\rm\bf updated\_occupancy\_grid}(z_{t},x_{t}^{[r]},m_{t-1}^{[r]})$
\State $\Theta_{t}^{[r]} = {E}[p(\Theta \mid x_{0:t}^{[r]}, \mathbf{C}_{1:t}^{[r]}, f_{1:t}, \mathbf{h})]$
\State $\bar{X}_{t} = \bar{X}_{t} \cup \langle x_{0:t}^{[r]}, \mathbf{C}_{1:t}^{[r]}, m_{t}^{[r]}, \Theta_{t}^{[r]}, \omega_{t}^{[r]} \rangle$
\EndFor
\State $S_{1:t}^{*} = \argmax_{S_{1:t}^{[r]}} \sum_{r=1}^{R} \omega_{t}^{[r]} \delta(S_{1:t} - S_{1:t}^{[r]})$
\State $LM_{t} = \argmax_{LM} p(LM \mid S_{1:t}^{*}, \lambda)$ 
\For{$r=1$ to $R$}
\State draw $i$ with probability $\propto \omega_{t}^{[i]}$ 
\State add $\langle x_{0:t}^{[i]}, \mathbf{C}_{1:t}^{[i]}, m_{t}^{[i]}, \Theta_{t}^{[i]}, LM_{t}  \rangle$ to $X_{t}$
\EndFor
\Return $X_{t}$
\EndProcedure
\end{algorithmic}
\end{algorithm}

\subsection{Online spatial concept acquisition and mapping}
Here, we describe the derivation of formulas for the online algorithm.
The online learning algorithm of the proposed method can be derived by introducing sequential update equations for estimating the parameters of the spatial concepts into the formulation of FastSLAM based on RBPF.
The proposed method assumes grid-based FastSLAM~2.0~\cite{gridbasedfastslam2005,gridbasedfastslam2007} algorithm.
Algorithm~\ref{alg:SpCoSLAM} is the online learning algorithm of SpCoSLAM.
As an advantage of using a particle filter, parallel processing can be easily applied because each particle can be calculated independently.

In the formulation of FastSLAM, the joint posterior distribution is factorized as follows:
\begin{eqnarray}
&&p(x_{0:t}, m \mid u_{1:t}, z_{1:t}) \nonumber \\
&&=\underbrace{p(m \mid x_{0:t}, z_{1:t})}_{\rm Mapping}\underbrace{p(x_{0:t} \mid u_{1:t}, z_{1:t})}_{\rm Particle~filter}.
\label{eq:fastslam}
\end{eqnarray}
This factorization represents a decomposition into two calculations: the mapping and self-localization by RBPF.

In the formulation of SpCoSLAM, the joint posterior distribution can be factorized to the probability distributions of a language model $LM$, a map $m$, the set of model parameters of spatial concepts $\Theta = \{ {\mathbf W}, \mbox{\boldmath $\mu $}, \mbox{\boldmath $\Sigma$}, {\mathbf \theta}, {\mathbf \phi}, \pi \}$, and the joint distribution of trajectory of self-position $x_{0:t}$ and the set of latent variables $\mathbf{C}_{1:t} = \{i_{1:t},C_{1:t},S_{1:t} \}$.
We describe the joint posterior distribution of SpCoSLAM as follows:
\begin{eqnarray}
&&p(x_{0:t},\mathbf{C}_{1:t}, LM, \Theta, m 
\mid u_{1:t}, z_{1:t}, y_{1:t}, f_{1:t}, AM ,\mathbf{h}) \nonumber \\
&&=p(LM \mid S_{1:t}, \lambda)p(m \mid x_{0:t}, z_{1:t}) \nonumber \\ 
&&\hspace{1.0em}\cdot~p(\Theta \mid x_{0:t}, \mathbf{C}_{1:t}, f_{1:t}, \mathbf{h}) \nonumber \\
&&\hspace{1.0em}\cdot~\underbrace{p(x_{0:t},\mathbf{C}_{1:t} \mid u_{1:t}, z_{1:t}, y_{1:t}, f_{1:t}, AM ,\mathbf{h})}_{\rm Particle~filter}
\label{eq:spcoslam}
\end{eqnarray}
where the set of hyperparameters is denoted as $\mathbf{h}= \{ \alpha,\beta,\gamma,\chi,\lambda, m_{0},\kappa_{0}, V_{0},\nu_{0} \}$.
Note that the speech signal $y_{t}$ is not observed at all times.
In this paper, the proposed method is equivalent to FastSLAM at the time when $y_{t}$ is not observed.

The particle filter algorithm uses sampling importance resampling (SIR).
We describe the importance weight $\omega_{t}^{[r]}$ for each particle as follows:
\begin{eqnarray}
\omega_{t}^{[r]}&=&\frac{p(x_{0:t}^{[r]},\mathbf{C}_{1:t}^{[r]} \mid u_{1:t}, z_{1:t}, y_{1:t}, f_{1:t}, AM ,\mathbf{h})}{q(x_{0:t}^{[r]},\mathbf{C}_{1:t}^{[r]} \mid u_{1:t}, z_{1:t}, y_{1:t}, f_{1:t}, AM ,\mathbf{h})} \nonumber \\
&=&\frac{P_{t}^{[r]}}{Q_{t}^{[r]}}
\label{eq:spcoslam_target_proposal}
\end{eqnarray}
where the particle index is $r$. The number of particles is $R$.
Henceforth, equations are also calculated for each particle $r$, but the subscripts representing the particle index are omitted.

We describe the target distribution $P_{t}$ as follows:
\begin{eqnarray}
&&p(x_{0:t},\mathbf{C}_{1:t} \mid u_{1:t}, z_{1:t}, y_{1:t}, f_{1:t}, AM ,\mathbf{h}) \nonumber \\
&&\approx p(z_{t} \mid x_{t}, m_{t-1})p(f_{t} \mid  C_{1:t-1}, f_{1:t-1}, \mathbf{h}) \nonumber \\
&&\hspace{1.0em}\cdot~p(i_{t},C_{t} \mid x_{0:t}, i_{1:t-1},C_{1:t-1},S_{1:t},f_{1:t},\mathbf{h}) \nonumber \\
&&\hspace{1.0em}\cdot~p(x_{t} \mid x_{t-1}, u_{t})p(S_{t} \mid S_{1:t-1},y_{1:t},AM,\lambda) \nonumber \\
&&\hspace{1.0em}\cdot~\frac{p(S_{t} \mid S_{1:t-1}, C_{1:t-1}, \alpha,\beta)}{p(S_{t} \mid S_{1:t-1}, \beta)} \cdot P_{t-1}. 
\label{eq:spcoslam_target}
\end{eqnarray}

We describe the proposal distribution $Q_{t}$ as follows:
\begin{eqnarray}
&&q(x_{0:t},\mathbf{C}_{1:t} \mid u_{1:t}, z_{1:t}, y_{1:t}, f_{1:t}, AM ,\mathbf{h}) \nonumber \\
&&= \underbrace{q(x_{t},\mathbf{C}_{t} \mid x_{0:t-1},\mathbf{C}_{1:t-1}, u_{1:t}, z_{1:t}, y_{1:t}, f_{1:t}, AM ,\mathbf{h})}_{q_{t}} \nonumber \\
&&\cdot\,\underbrace{q(x_{0:t-1},\mathbf{C}_{1:t-1} | u_{1:t-1}, z_{1:t-1}, y_{1:t-1}, f_{1:t-1}, AM ,\mathbf{h})}_{Q_{t-1}} \nonumber \\
&&=q_{t}Q_{t-1}.
\label{eq:spcoslam_proposal}
\end{eqnarray}

The weight $\omega_{t}$ is represented by (\ref{eq:spcoslam_target_proposal}), (\ref{eq:spcoslam_target}), and (\ref{eq:spcoslam_proposal}) as follows: 
\begin{eqnarray}
\omega_{t} 
&\approx& p(z_{t} \mid x_{t}, m_{t-1})p(f_{t} \mid  C_{1:t-1}, f_{1:t-1}, \mathbf{h}) \nonumber \\
&&\cdot~p(i_{t},C_{t} \mid x_{0:t}, i_{1:t-1},C_{1:t-1},S_{1:t},f_{1:t},\mathbf{h}) \nonumber \\
&&\cdot~p(x_{t} \mid x_{t-1}, u_{t})p(S_{t} \mid S_{1:t-1},y_{1:t},AM,\lambda) \nonumber \\
&&\cdot~\frac{p(S_{t} \mid S_{1:t-1}, C_{1:t-1}, \alpha,\beta)}{p(S_{t} \mid S_{1:t-1}, \beta) q_{t}} 
\cdot \underbrace{\frac{P_{t-1}}{Q_{t-1}}}_{\omega_{t-1}}. 
\label{eq:spcoslam_target_proposal2}
\end{eqnarray}

We assume the proposal distribution $q_{t}$ at time $t$ as follows:
\begin{eqnarray}
q_{t}
&=&p(x_{t} \mid x_{t-1},z_{t},m_{t-1},u_{t}) \nonumber \\
&&\cdot~p(i_{t},C_{t} \mid x_{0:t}, i_{1:t-1},C_{1:t-1},S_{1:t},f_{1:t},\mathbf{h}) \nonumber \\
&&\cdot~p(S_{t} \mid S_{1:t-1},y_{1:t},AM,\lambda). 
\label{eq:spcoslam_proposal2}
\end{eqnarray}

Then, $p(x_{t} \mid x_{t-1},z_{t},m_{t-1},u_{t})$ is equivalent to the proposal distribution of FastSLAM~2.0.

The term of $i_{t}$ and $C_{t}$ is the marginal distribution regarding the set of model parameters $\Theta$.
This distribution can be calculated by a formula equivalent to collapsed Gibbs sampling.
We describe the equation for sampling $i_{t}$ and $C_{t}$ simultaneously as follows:
\begin{eqnarray}
&&p(i_{t},C_{t} \mid x_{0:t}, i_{1:t-1},C_{1:t-1},S_{1:t},f_{1:t},\mathbf{h}) \nonumber \\
&&\propto p(S_{1:t} \mid C_{1:t}, \beta)p(f_{1:t} \mid C_{1:t}, \chi)p(x_{0:t} \mid i_{1:t}, \mathbf{h}) \nonumber \\
&&\hspace{1.0em}\cdot~p(i_{t}, C_{t} \mid i_{1:t-1}, C_{1:t-1}, \alpha, \gamma). 
\label{eq:spcoslam_itct}
\end{eqnarray}
The details of (\ref{eq:spcoslam_itct}) are described in Section~\ref{sec:SpCoSLAM:sampling}.

We approximate the term of $S_{t}$ by speech recognition using the language model $LM_{t-1}$ and unsupervised word segmentation using the WFST speech recognition results ${\cal L}_{1:t}$ as follows:
\begin{eqnarray}
&&\hspace{-1.2em}p(S_{t} \mid S_{1:t-1},y_{1:t},AM,\lambda) \nonumber \\
&&\hspace{-1.2em}\approx latticelm(S_{1:t} \mid {\cal L}_{1:t},\lambda){\rm SR}({\cal L}_{1:t} \mid y_{1:t},AM,LM_{t-1}). \nonumber \\
\label{eq:spcoslam_latticelm_SR}
\end{eqnarray}
In the formulation of (\ref{eq:spcoslam}), it is desirable to estimate the language model $LM_{t}$ for each particle.
However, in this case, it is necessary to perform speech recognition of the number of data times the number of particles for each teaching utterance.
In order to reduce the computational cost, we use a language model $LM_{t}$ of a particle with the maximum weight for speech recognition of the next step.

Finally, $\omega_{t}$ is represented as follows:
\begin{eqnarray}
\omega_{t} 
&\approx& p(z_{t} \mid m_{t-1}, x_{t-1},u_{t})p(f_{t} \mid  C_{1:t-1}, f_{1:t-1}, \mathbf{h}) \nonumber \\
&&\cdot~\frac{p(S_{t} \mid S_{1:t-1}, C_{1:t-1}, \alpha,\beta)}{p(S_{t} \mid S_{1:t-1}, \beta)} 
\cdot \omega_{t-1}. 
\label{eq:spcoslam_weight}
\end{eqnarray}
This is an equation obtained by multiplying the weight $\omega_{t-1}$ at a previous time with the marginal likelihoods for $z_{t}$, $f_{t}$, and $S_{t}$.

\subsection{Simultaneous sampling of indices $i_{t}$ and $C_{t}$}
\label{sec:SpCoSLAM:sampling} 
The proposed method uses the Chinese restaurant process (CPR)~\cite{aldous1985exchangeability}, which is one of the constitution methods of the Dirichlet process (DP). 
We describe the distribution of $C_{t}$ using the CRP representation as follows:
\begin{eqnarray}
p(C_{t}=l \mid C_{1:t-1}, \alpha) = 
\left\{ \begin{array}{ll}
    \frac{n^{(l)}_{t}}{n_{t} + \alpha} & (n^{(l)}_{t} > 0) \\
    \frac{\alpha}{n_{t} + \alpha} & (l~{\rm is~new})
  \end{array} \right.
\label{eq:prior_ct}
\end{eqnarray}
where $n^{(l)}_{t}$ denotes the number of data allocated to the $l$-th spatial concept in all data up to the time $t-1$.
The number of data is $n_{t} = \sum_{l'}n^{(l')}_{t}$.

We describe the distribution of $i_{t}$ by the CRP representation as follows:
\begin{eqnarray}
&&p(i_{t}=k \mid i_{1:t-1}, C_{1:t-1}, C_{t} = l, \gamma) \nonumber \\
&&\hspace{0.0em}= \left\{ \begin{array}{ll}
    \frac{n^{(l,k)}_{t}}{n^{(l)}_{t} + \gamma} & (n^{(l,k)}_{t} > 0) \\
    \frac{\gamma}{n^{(l)}_{t} + \gamma} & (k~{\rm is~new})
  \end{array} \right.
\label{eq:prior_it}
\end{eqnarray}
where $n^{(l,k)}_{t}$ denotes the number of data allocated to the $k$-th position distribution in data allocated to the $l$-th spatial concept. 

Therefore, the joint prior distribution of $i_{t}$ and $C_{t}$ is represented as follows:
\begin{eqnarray}
&&p(i_{t}=k, C_{t}=l \mid i_{1:t-1}, C_{1:t-1}, \alpha, \gamma) \nonumber \\
&&\hspace{0.0em}= \left\{ \begin{array}{ll}
    \frac{n^{(l,k)}_{t}}{n^{(l)}_{t} + \gamma} \frac{n^{(l)}_{t}}{n_{t} + \alpha} & (n^{(l,k)}_{t} > 0) \\
    \frac{\gamma}{n^{(l)}_{t} + \gamma} \frac{n^{(l)}_{t}}{n_{t} + \alpha} & (n^{(l)}_{t} > 0~\cap~k~{\rm is~new}) \\
    \frac{\gamma}{n^{(l)}_{t} + \gamma} \frac{\alpha}{n_{t} + \alpha} & (l~{\rm and}~k~{\rm are~new})
  \end{array} \right.
\label{eq:prior_itct}
\end{eqnarray}

The probability of words $S_{t}$ is represented as follows:
\begin{eqnarray}
&&p(S_{1:t} \mid C_{1:t-1}, C_{t}=l, \beta) \nonumber \\
&&= \prod_{B_{t}} p(S_{t,b}=s_{g}, S_{1:t-1} \mid C_{1:t-1}, C_{t}=l, \beta) \nonumber \\
&&\propto \left\{ \begin{array}{ll}
    \prod_{B_{t}} \frac{n^{(l,g)}_{t} + \beta}{\sum_{g'=1}^{G} (n^{(l,g')}_{t} + \beta)} & (n^{(l)}_{t} > 0) \\
    \frac{1}{G^{B_{t}}} & (l~{\rm is~new})
  \end{array} \right.
  \label{eq:prior_st}
\end{eqnarray}
where $G$ denotes the number of types of words, i.e., the number of dimensions of the multinomial distribution of the names of places and $n^{(l,g)}_{t}$ denotes the total number of words $s_{g}$ of the $g$-th dimension allocated to the $l$-th multinomial distribution of the names of the places in words $S_{1:t-1}$.

The probability of image features $f_{t}$ is represented as follows:
\begin{eqnarray}
&&p(f_{1:t} \mid C_{1:t-1}, C_{t}=l, \chi) \nonumber \\
&&= \prod_{E} p(f_{t,e}, f_{1:t-1} \mid C_{1:t-1}, C_{t}=l, \chi) \nonumber \\
&&\propto \left\{ \begin{array}{ll}
    \prod_{E} \Bigl( \frac{n^{(l,e)}_{t} + \chi}{\sum_{e'=1}^{E} (n^{(l,e')}_{t} + \chi)} \Bigr)^{f_{t,e}} & (n^{(l)}_{t} > 0) \\
    \frac{1}{E^{F_{t}}} & (l~{\rm is~new})
  \end{array} \right.
\label{eq:prior_ft}
\end{eqnarray}
where $E$ denotes the number of dimensions of image features, $n^{(l,e)}_{t}$ denotes the total number of image features of the $e$-th dimension 
allocated to the $l$-th multinomial distribution of image features in image features $f_{1:t-1}$, and  
$F_{t}=\sum_{E} f_{t,e}$.

The probability of self-position $x_{t}$ of the robot is described as follows: 
\begin{eqnarray}
&&p(x_{t}, x_{0:t-1} \mid i_{1:t-1},i_{t}=k, \mathbf{h}) \nonumber \\
&&\propto {\rm St}(x_{t} \mid m_{k}, \frac{V_{q}(\kappa_{k} + 1)}{\kappa_{k}(\nu_{k}-d+1)}, \nu_{k}-d+1) 
\label{eq:prior_xt}
\end{eqnarray}
where the function ${\rm St}()$ denotes the multivariate Student's t-distribution~\cite{murphy2012machine}.
Then, the posterior parameters in (\ref{eq:prior_xt}) are represented as follows:
\begin{eqnarray}
\bar{x}_{k} &=& \frac{1}{n_{t}^{(k)}} \sum_{x_{j} \in {\bf x}_{k}} x_{j} \\
m_{k} &=& \frac{n_{t}^{(k)} \bar{x}_{k} + \kappa_{0} m_{0}}{n_{t}^{(k)} + \kappa_{0}} \\
\kappa_{k} &=& n_{t}^{(k)}+\kappa_{0} \\
\nu_{k} &=& \nu_{0}+n_{t}^{(k)} \\
V_{q} &=& V_{0} + \sum_{x_{j} \in {\bf x}_{k}} x_{j}x_{j}^{\rm T} + \kappa_{0} m_{0}m_{0}^{\rm T} - \kappa_{k} m_{k}m_{k}^{\rm T} 
\end{eqnarray}
where $n^{(k)}_{t}$ and ${\bf x}_{k}$ are the number of data and the set of position data, respectively, allocated to the position distribution of $i_{t}=k$ in data up to the time $t-1$.

From the above, (\ref{eq:spcoslam_itct}) can be expressed as follows:
\begin{eqnarray}
&&\hspace{-1.0em}p(i_{t}=k,C_{t}=l \mid x_{0:t}, i_{1:t-1},C_{1:t-1},S_{1:t},f_{1:t},\mathbf{h}) \nonumber \\
&&\hspace{0.0em}\propto \left\{ \begin{array}{l}
    \prod_{B_{t}} \frac{n^{(l,g)}_{t} + \beta}{\sum_{g'=1}^{G} (n^{(l,g')}_{t} + \beta)} \prod_{E} \Bigl( \frac{n^{(l,e)}_{t} + \chi}{\sum_{e'=1}^{E} (n^{(l,e')}_{t} + \chi)} \Bigr)^{f_{t,e}} \\
    \cdot {\rm St}(x_{t} \mid m_{k}, \frac{V_{q}^{-1}(\kappa_{k} + 1)}{\kappa_{k}(\nu_{k}-d+1)}, \nu_{k}-d+1) \\
    \cdot \frac{n^{(l,k)}_{t}}{n^{(l)}_{t} + \gamma} \frac{n^{(l)}_{t}}{n_{t} + \alpha} \\
    \hspace{9.0em}(n^{(l,k)}_{t} > 0) \\
    \prod_{B_{t}} \frac{n^{(l,g)}_{t} + \beta}{\sum_{g'=1}^{G} (n^{(l,g')}_{t} + \beta)} \prod_{E} \Bigl( \frac{n^{(l,e)}_{t} + \chi}{\sum_{e'=1}^{E} (n^{(l,e')}_{t} + \chi)} \Bigr)^{f_{t,e}} \\
    \cdot {\rm St}(x_{t} \mid m_{0}, \frac{V_{0}^{-1}(\kappa_{0} + 1)}{\kappa_{0}(\nu_{0}-d+1)}, \nu_{0}-d+1) \\
    \cdot \frac{\gamma}{n^{(l)}_{t} + \gamma} \frac{n^{(l)}_{t}}{n_{t} + \alpha} \\
    \hspace{9.0em}(n^{(l)}_{t} > 0~\cap~k~{\rm is~new}) \\
    \frac{1}{G^{B_{t}}} \frac{1}{E^{F_{t}}} \\
    \cdot {\rm St}(x_{t} \mid m_{0}, \frac{V_{0}^{-1}(\kappa_{0} + 1)}{\kappa_{0}(\nu_{0}-d+1)}, \nu_{0}-d+1) \\
    \cdot \frac{\gamma}{n^{(l)}_{t} + \gamma} \frac{\alpha}{n_{t} + \alpha} \\
    \hspace{9.0em}(l~{\rm and}~k~{\rm are~new})
  \end{array} \right.
\label{eq:spcoslam_itct2}
\nonumber \\
\end{eqnarray}

\section{Experiments}
We performed experiments for online learning of spatial concepts from a novel environment.
In addition, we performed evaluations of place categorization and lexical acquisition related to place.
We compare the performance of four methods as follows: 
\begin{enumerate}
\item[(A)] SpCoSLAM 
\item[(B)] Online SpCoA based on RBPF 
\item[(C)] Online SpCoA 
\item[(D)] SpCoA (Batch learning)~\cite{taniguchi_spcoa}
\end{enumerate}

Methods (A), (B), and (C) performed online learning algorithms based on the CRP representation. 
Methods (B), (C), and (D) based on SpCoA did not perform the update of a language model and did not use image features.
Method (D) performed Gibbs sampling based on a weak-limit approximation~\cite{fox2011sticky} of the stick-breaking process (SBP)~\cite{sethuraman1994constructive}, i.e., the upper limit numbers of spatial concepts and position distributions were set as $L=100$ and $K=100$ respectively. %
In the batch learning (D), we performed Gibbs sampling for 100 iterations.

\begin{figure*}[tb]
  \begin{center}
    \includegraphics[width=0.9\hsize]{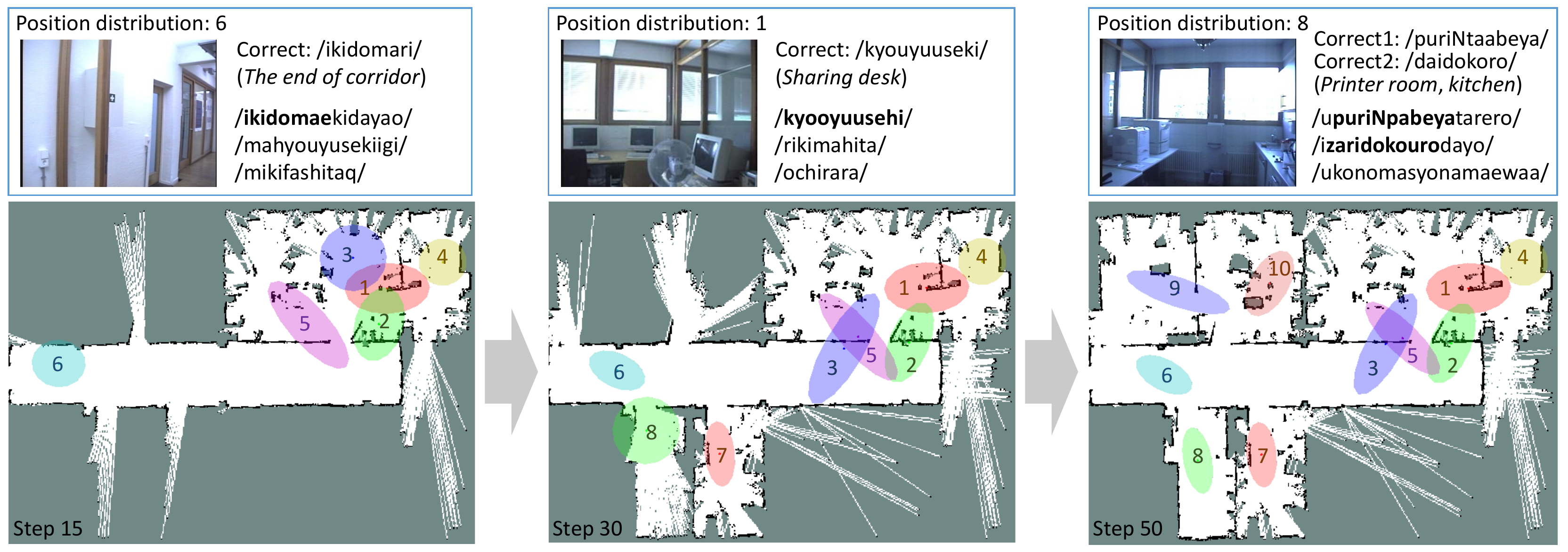}
    \caption{Learning results of each position distribution in a generated map; Ellipses denoting the position distributions are drawn on the map at steps 15, 30, and 50.
The colors of the ellipses were determined randomly. Furthermore, each index number is denoted as $i_{t}=k$.}
    \label{fig:map}
  \end{center}
\end{figure*}

\subsection{Online learning}
We conducted experiments for online spatial concept acquisition in a real environment. 
We extended the gmapping package, implementing the grid-based FastSLAM~2.0~\cite{gridbasedfastslam2005,gridbasedfastslam2007} in the robot operating system (ROS). 
We used an open dataset (albert-b-laser-vision) containing a rosbag file in which the odometry, laser range data, and vision data were recorded.
This dataset was obtained from the Robotics Data Set Repository (Radish)~\cite{Radish}. The authors thank Cyrill Stachniss for providing this data.
We prepared a Japanese speech signal data corresponding to the movement of the robot of the above dataset because it did not include speech signal data.
The number of teaching places was 10 and there were nine place names.
The teaching utterances included 10 types of various phrases.
The total number of utterances was 50. 
The employed microphone was a SHURE PG27-USB.
The speech recognition system uses Julius dictation-kit-v4.3.1-linux (GMM-HMM decoding)~\cite{kawahara1998sharable}.
The initial word dictionary of the Julius system contains 115 Japanese syllables.
The unsupervised word segmentation system uses latticelm~\cite{neubig2012bayesian}.
We used a deep learning framework Caffe~\cite{jia2014caffe} for CNNs as an image feature extractor.
We used a pre-trained CNN, i.e., Places205-AlexNet trained on 205 scene categories of Places Database with $2.5\times 10^6$ images~\cite{zhou2014learning}.
The map resolution was 0.05~m/grid.
The number of particles was $R=30$.
The hyperparameters were set as follows: $\alpha =20$, $\gamma =10$, $\beta=0.2$, $\chi=0.2$, $m_{0}=[ 0 , 0 ]^{\rm T}$, $\kappa_{0}=0.001$, $V_{0}={\rm diag}(2,2)$, and $\nu_{0}=3$. 
The above parameters were set so that all methods in the comparison were tested under the same conditions.

%
Fig.~\ref{fig:map} shows the position distributions in the environmental maps at steps 15, 30, and 50.
The upper part of this figure shows an example of the image corresponding to each position distribution, the correct phoneme sequence of the name of the place, and the upper three words of the probability value estimated by the probability distribution $p(S_{t} \mid i_{t}, \Theta_{t}, LM_{t})$ at step $t$.
As a result, Fig.~\ref{fig:map} shows how the spatial concepts are acquired while sequentially mapping.
Details on online learning experiment can be seen in the video attachment.

\subsection{Estimation accuracy of spatial concepts} 
We compare the matching rate for the estimated index $C_{t}$ of the spatial concept of each teaching utterance and the classification results of correct answers by a person.
In this experiment, the evaluation metric uses the normalized mutual information (NMI), which is a measure of the degree of similarity between two clustering results.
The estimated index $i_{t}$ of the position distributions is also evaluated in the same manner.
In addition, we evaluate the estimated number of spatial concepts $L$ and position distributions $K$ by using the estimation accuracy rate (EAR).
The EAR was calculated as follows:
\begin{eqnarray}
{\rm EAR}= {\rm min}( 1 - \frac{\mid n_{\rm T} - n_{\rm E} \mid}{n_{\rm T}}, 0 )
 \label{eq:EAR}
\end{eqnarray}
where $n_{\rm T}$ is the correct number and $n_{\rm E}$ is the estimated number.

\begin{figure}[tb]
 \begin{minipage}{0.495\hsize}
  \begin{center}
    \includegraphics[width=1.1\hsize]{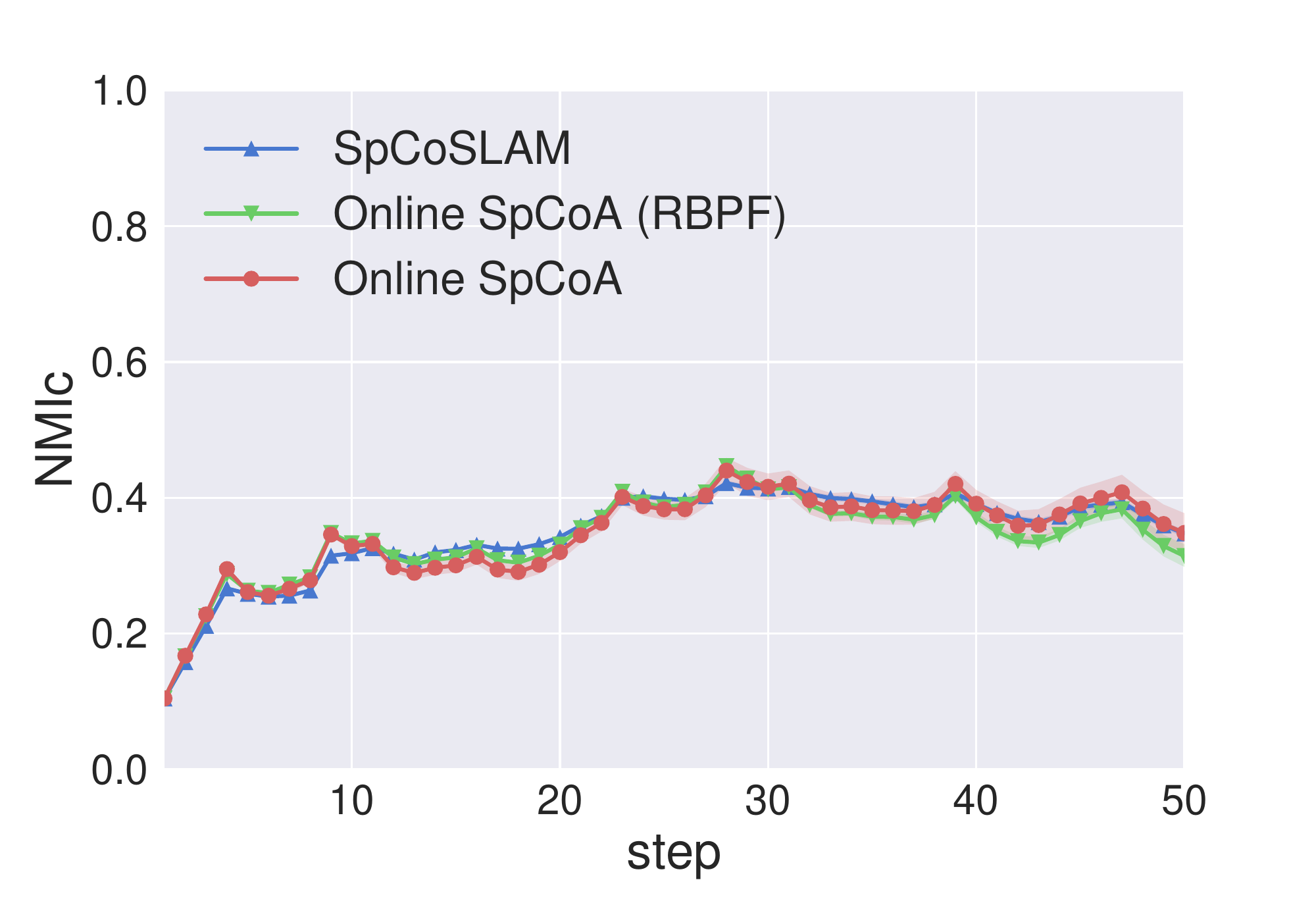}
    {\footnotesize (a) NMI values of $C_{1:t}$}
  \end{center}
     \end{minipage}
 \begin{minipage}{0.495\hsize}
  \begin{center}
    \includegraphics[width=1.1\hsize]{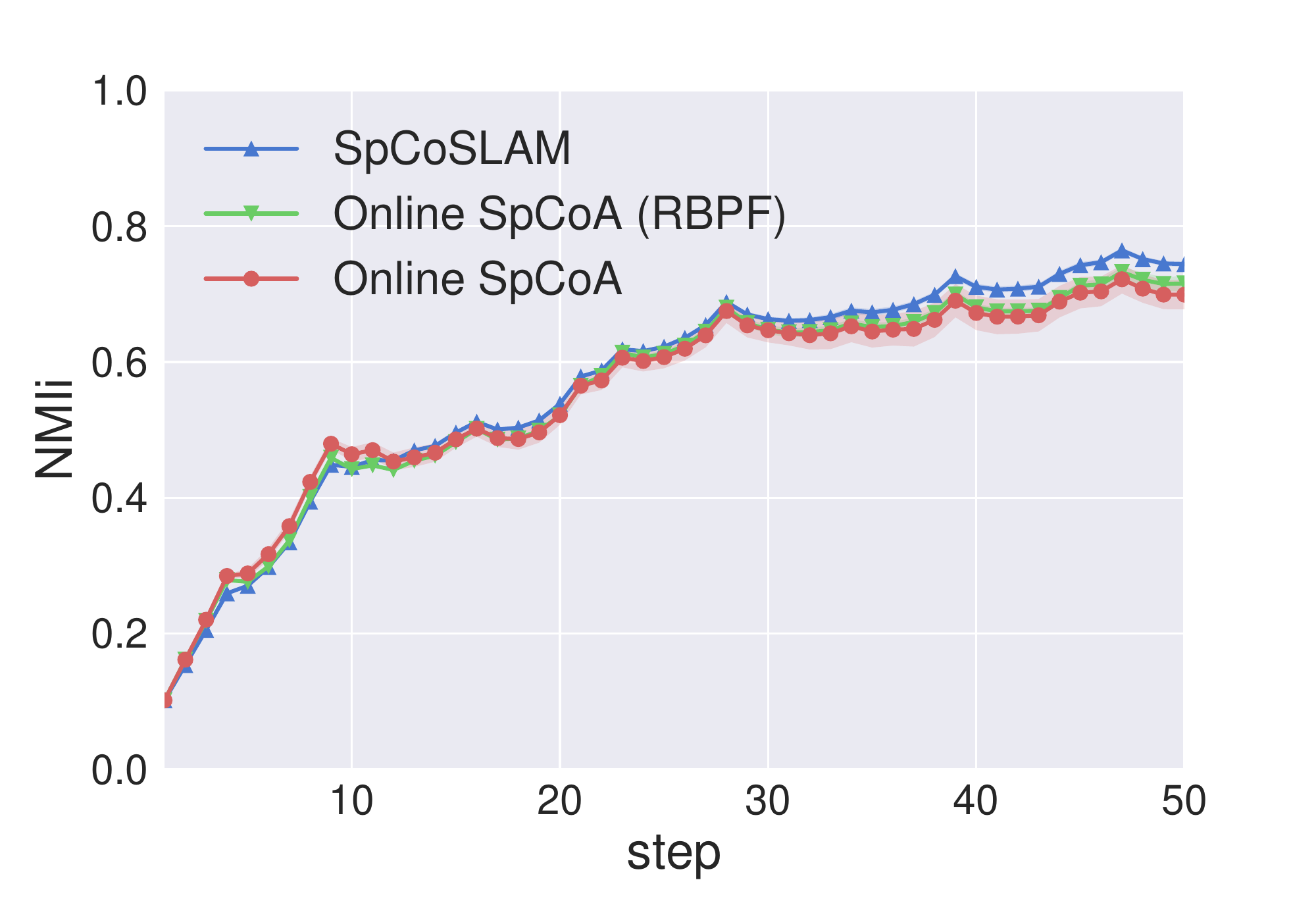}
    {\footnotesize (b) NMI values of $i_{1:t}$}
  \end{center}
       \end{minipage}
       \caption{Accuracy rates of the estimation results for (a) index of spatial concepts $C_{1:t}$ and (b) index of position distributions $i_{1:t}$}
    \label{fig:NMIic}
\end{figure}

\begin{figure}[tb]
 \begin{minipage}{0.495\hsize}
  \begin{center}
    \includegraphics[width=1.1\hsize]{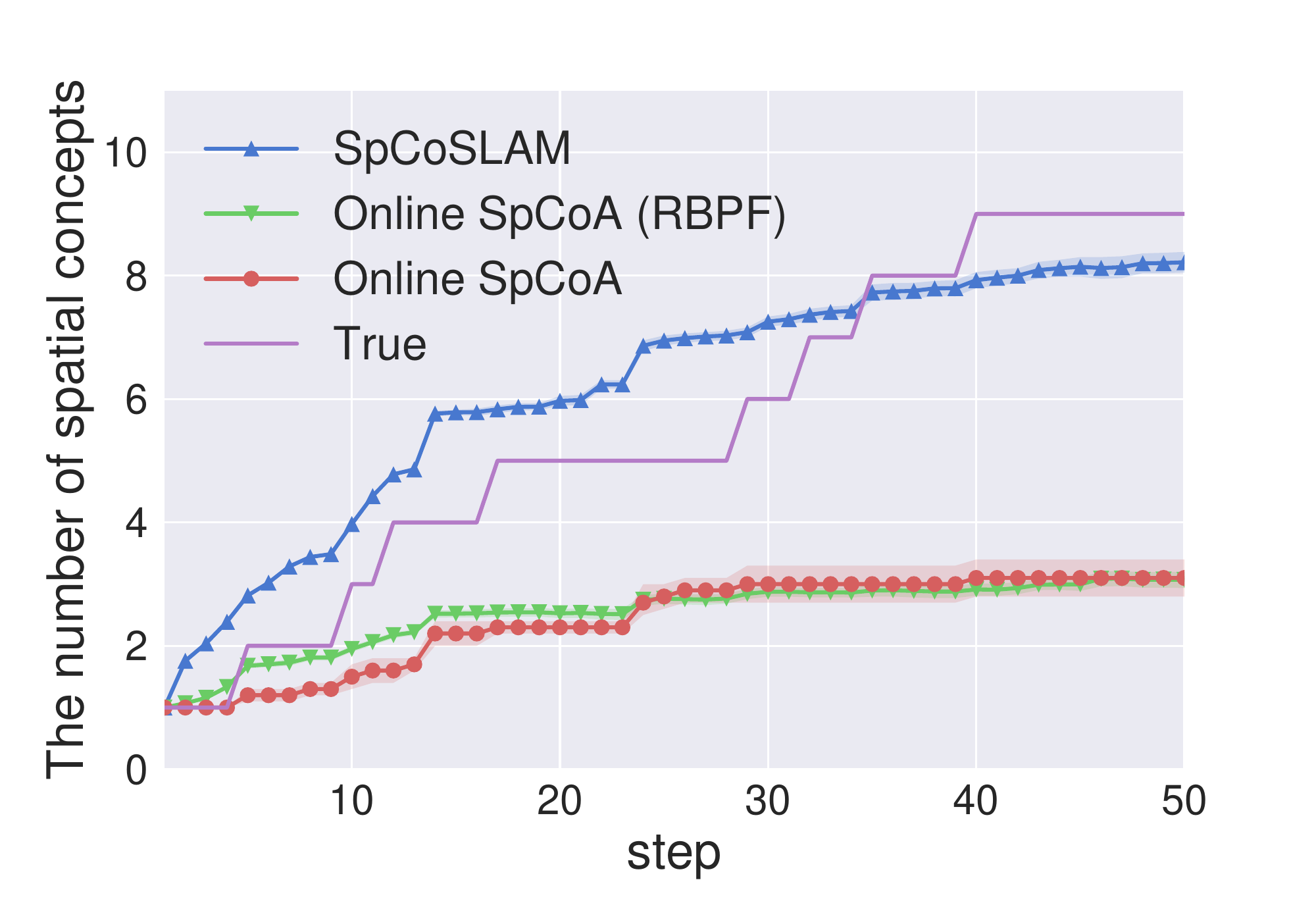}
    {\footnotesize (a) Number of spatial concepts} 
  \end{center}
     \end{minipage}
 \begin{minipage}{0.495\hsize}
  \begin{center}
    \includegraphics[width=1.1\hsize]{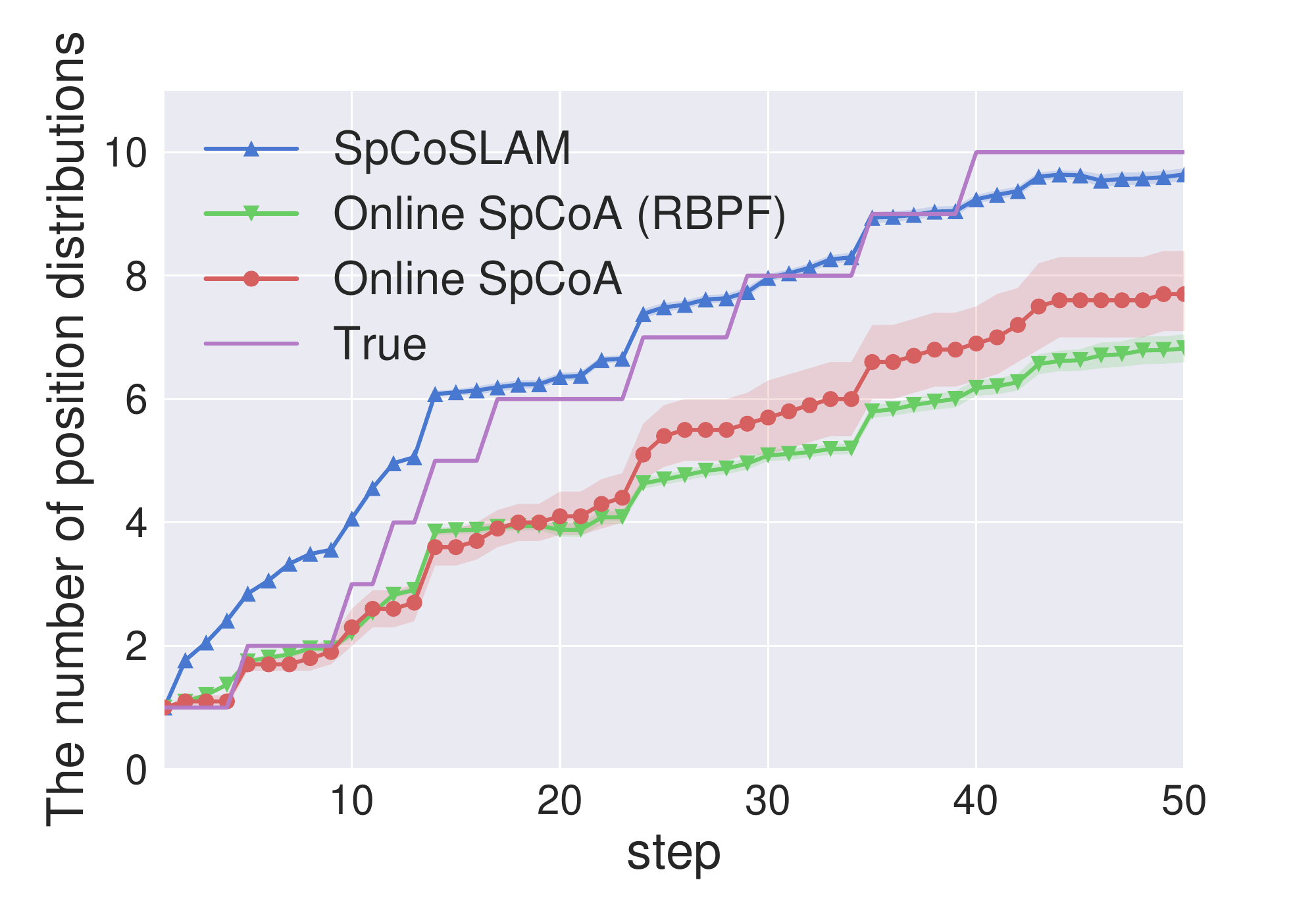}
    {\footnotesize (b) Number of position distributions} 
  \end{center}
       \end{minipage}
       \caption{Estimation results for (a) the number of spatial concepts $L$ and (b) the number of position distributions $K$}
    \label{fig:LK}
\end{figure}

Table~\ref{hyouka2} lists the evaluation-value averages calculated using the metrics NMI and EAR at step 50.
Fig.~\ref{fig:NMIic} shows the average of the NMI values in 10 trials by online learning.
In both $C_{t}$ and $i_{t}$, the NMI values tended to rise at the beginning.
The NMI values of $C_{t}$ were similar for methods (A), (B), and (C).
In the NMI values for $i_{t}$, the proposed method (A) showed higher values than the other methods after step 30.
We consider a major possible reason for the clustering results of spatial concepts.
In online lexical acquisition, the word segmentation results cannot be obtained stably when training dataset is small.
We consider that stable words can be obtained by further increasing the number of training steps.
Fig.~\ref{fig:LK} shows the average of the number of spatial concepts and the number of position distributions in 10 trials by online learning.
The average values of the estimated results of method~(D) were $L = 18.9$, $K = 13.1$.
True data was determined by a user based on teaching data.
The experimental results show that the proposed method~(A) was closer to the true data than other methods for both $L$ and $K$. 


\begin{table}[tb] 
\begin{center}
\caption{Evaluation values of NMI and EAR for each method}
\begin{tabular}{p{70pt}|cc|cc} \hline
 & \multicolumn{2}{|c|}{NMI} & \multicolumn{2}{|c}{EAR} \\  
Methods & $C_{t}$ & $i_{t}$ & $L$ & $K$ \\ \hline
(A) SpCoSLAM & 0.347 & \underline{0.744} & \underline{\bf 0.913} & \underline{\bf 0.964} \\ 
(B) Online SpCoA (RBPF) & {\shortstack{\raisebox{-5pt}{0.314}}} & {\shortstack{\raisebox{-5pt}{0.716}}} & {\shortstack{\raisebox{-5pt}{0.341}}} & {\shortstack{\raisebox{-5pt}{0.682}}} \\ 
(C) Online SpCoA & \underline{0.348} & 0.699 & \underline{0.344} & 0.770 \\ 
(D) SpCoA~\cite{taniguchi_spcoa} & \underline{\bf 0.805} & \underline{\bf 0.856} & 0.000 & \underline{0.690} \\ \hline 
\end{tabular}
\label{hyouka2}
\end{center}
\end{table}

\subsection{Comparison of the number of segmented words}
\begin{figure}[tb]
  \begin{center}
    \includegraphics[width=185pt]{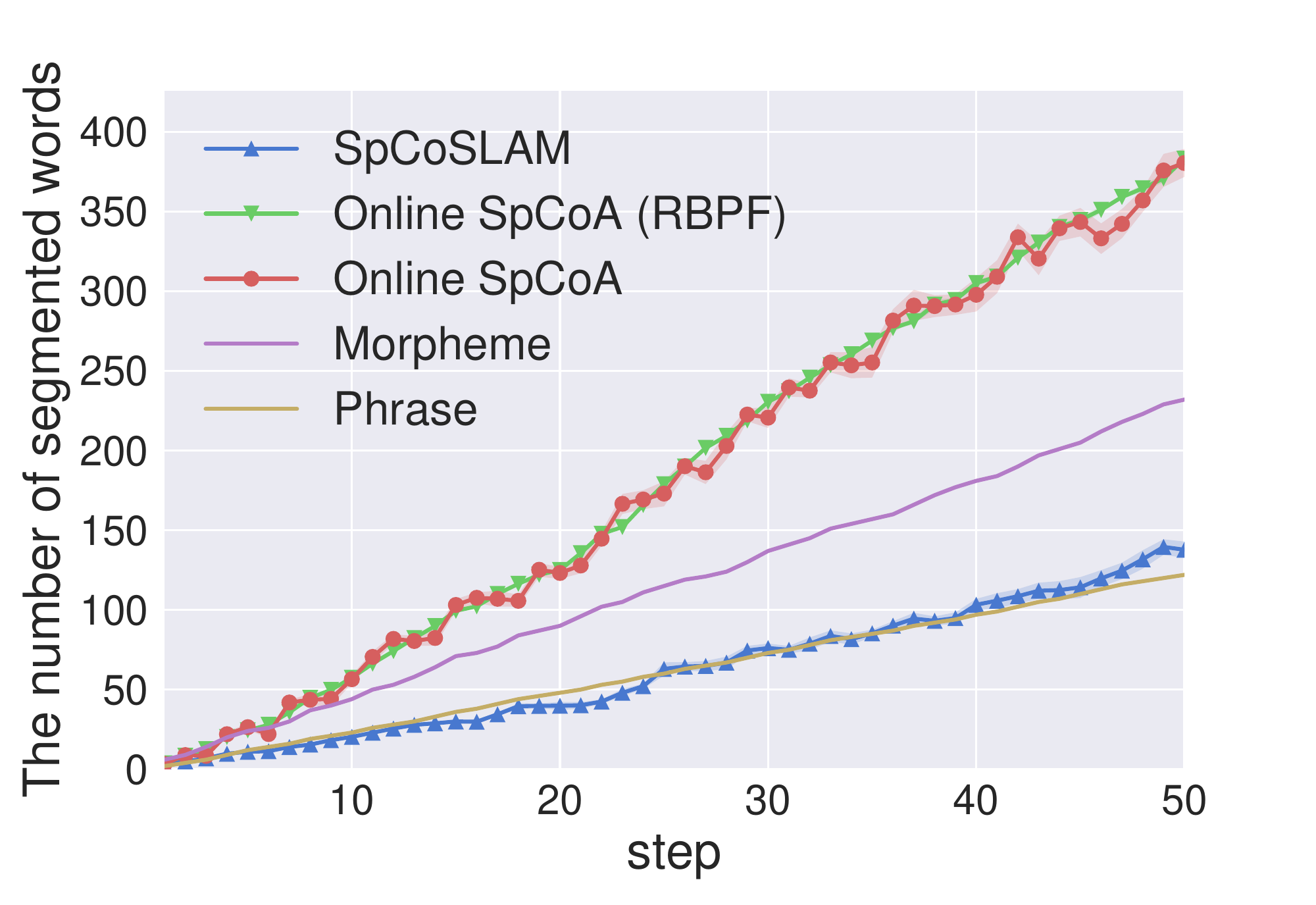}
    \caption{Number of segmented words}
    \label{fig:seg}
  \end{center}
\end{figure}

We show whether a phoneme sequence including the name of a place is properly segmented.
Fig.~\ref{fig:seg} shows the number of segmented words.
The morphological segmentation (purple line) was suitably segmented into Japanese morphemes using MeCab, which is an off-the-shelf Japanese morphological analyzer that is widely used for natural language processing. 
The phrase segmentation (yellow line) was the number of words in the case of segmenting words only before and after the name of the place, i.e., we assume that a phrase other than the name of the place is one word. 
Table~\ref{bunkatu} presents examples of the word segmentation results of the four methods.
Method (A) was similar to the phrase segmentation.
On the other hand, methods (B) and (C) showed results of over-segmentation.
In addition, the average value of the number of segmented words of method (D) was 391.4, i.e., it was similar to methods (B) and (C) at step 50.
The results indicate that method (A) improved the problem of over-segmentation by updating the language model sequentially.

\begin{table*}[tb]
\renewcommand{\arraystretch}{1.3}
\begin{center}
\caption{Examples of word segmentation results of uttered sentences. {\it ``$\mid$''} denotes a word segment position}
\begin{tabular}{c||c|c|c} \hline
English & 
{\it ``We come to {\bf the end of corridor}.''} & 
{\it ``{\bf The faculty laboratory} is here.''} & 
{\it ``This place name is {\bf the break room}.''} 
\\ \hline \hline
Morpheme & 
{\bf ikidomari}$\mid$ni$\mid$ki$\mid$mashi$\mid$ta & 
{\bf kyouiNkeNkyuushitsu}$\mid$wa$\mid$kochira$\mid$desu & 
kono$\mid$basyo$\mid$no$\mid$namae$\mid$wa$\mid${\bf kyuukeijo} \\ \hline
Phrase & 
{\bf ikidomari}$\mid$nikimashita & 
{\bf kyouiNkeNkyuushitsu}$\mid$wakochiradesu & 
konobasyononamaewa$\mid${\bf kyuukeijo} \\ \hline
(A) & 
{\bf aaerikidomari}$\mid$nikeiwasuta & 
{\bf kyoiiNiNteNkyushitsu}$\mid$waqgochigadesu & 
ukonomasyonamaewaa$\mid${\bf kyuuqkirijo} \\ \hline
(B), (C), (D) & 
{\bf pikido$\mid$ma$\mid$e}$\mid$ni$\mid$ki$\mid$ma$\mid$sya   & 
{\bf kyoo$\mid$i$\mid$N$\mid$teN$\mid$kyu$\mid$shi$\mid$su}$\mid$wa$\mid$ko$\mid$chi$\mid$ga$\mid$desu & 
kono$\mid$basyo$\mid$no$\mid$nama$\mid$e$\mid$wa$\mid${\bf kyuu$\mid$ke$\mid$i$\mid$jo} \\ \hline
\end{tabular}
\label{bunkatu}
\end{center}
\end{table*}

\subsection{Place recognition using a speech signal}
When the robot hears a user's speech signal $y_{t}$ including the name of a place, the robot estimates a position $x_{t}^{\rm (best)}$ indicated by the uttered sentence.
The user says {\it ``** ni iqte.''} (which means {\it ``Go to **.''} in English).
The estimation of a position was calculated as follows:
\begin{eqnarray}
x_{t}^{\rm (best)} &=& \argmax_{x_{t}} p(x_{t} \mid y_{t}, \Theta, AM, LM). 
 \label{eq:xbest0}
\end{eqnarray}
In this experiment, (\ref{eq:xbest0}) was approximated by using the speech recognition results $S_{t}^{\rm (1:10\mathchar`-best)}$ from 1-best to 10-best as follows:
\begin{eqnarray}
S_{t}^{\rm (1:10\mathchar`-best)} &\sim& {\rm SR}(S_{t} \mid y_{t}, AM, LM), \\ 
 \label{eq:1best2}
x_{t}^{\rm (best)} &=& \argmax_{x_{t}} p(x_{t} \mid S_{t}^{\rm (1:10\mathchar`-best)}, \Theta).
 \label{eq:xbest}
\end{eqnarray}
It is difficult to calculate (\ref{eq:xbest}) for all of the possible positions.
Therefore, we use 10 position coordinates sampled for each position distribution as candidates for $x_{t}^{\rm (best)}$.
As a justification for this, we consider that positions near the mean values of position distributions become possible candidates for calculating (\ref{eq:xbest}). 
In this experiment, we decided to correct the position within the rectangular area surrounding the position coordinates taught as the same place (including 0.5~m margins to the right, left, above, and below).
The place recognition rate (PRR) is calculated as follows:
\begin{eqnarray}
{\rm PRR} =  \frac{n_{\rm C}}{n_{\rm U}},
 \label{eq:PRR}
\end{eqnarray}
where $n_{\rm U}$ denotes the number of utterances and $n_{\rm C}$ denotes the number of correct positions.
The number of utterances is nine. 

\begin{figure}[tb]
  \begin{center}
    \includegraphics[width=185pt]{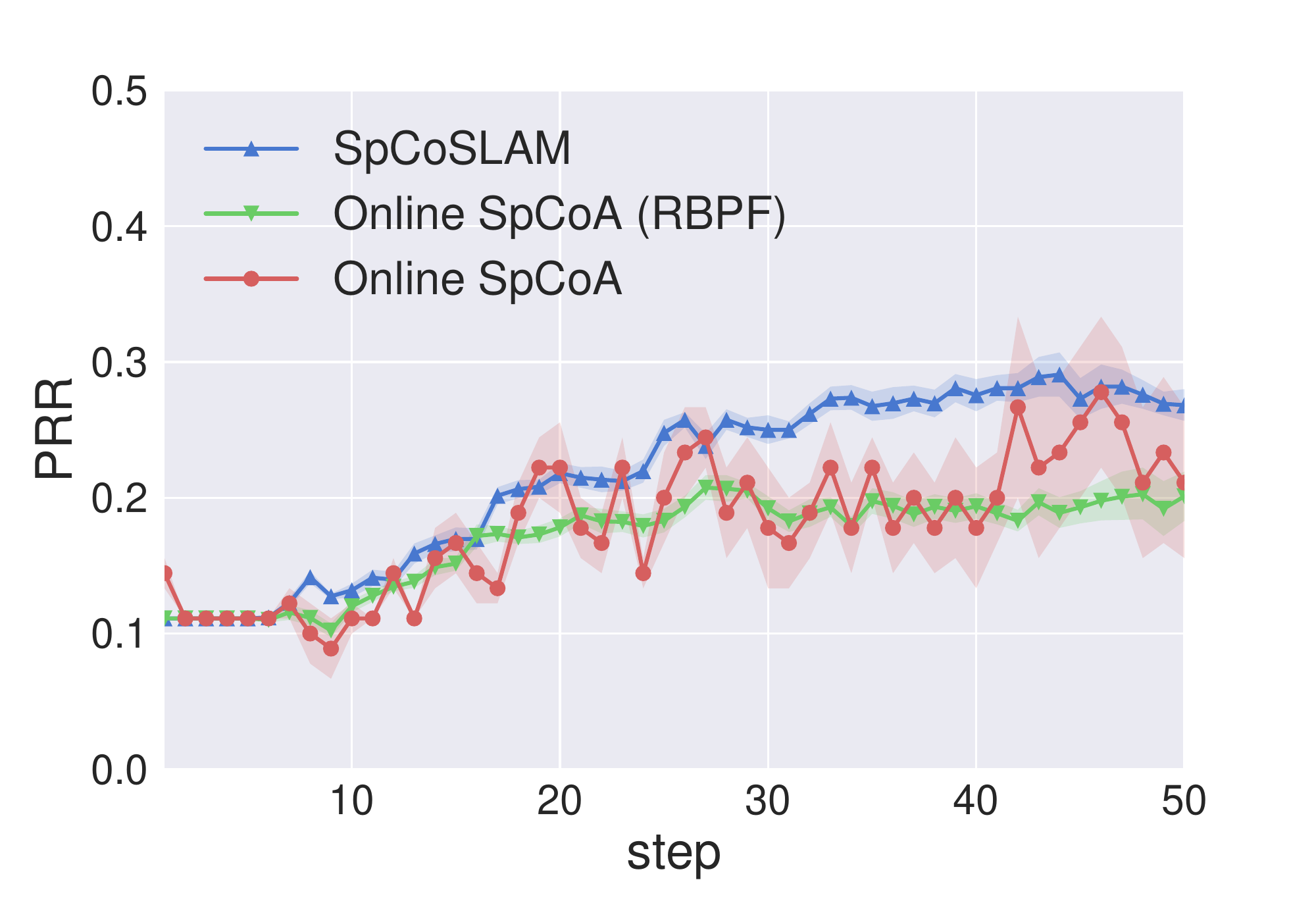}
    \caption{Results of PRR values}
    \label{fig:PRR}
  \end{center}
\end{figure}

Fig.~\ref{fig:PRR} shows the average of the PRR values in 10 trials.
The average value of PRR of Method (D) was 0.500.
Method (A) showed the highest overall evaluation values of the online methods.
The experimental results show that the robot was able to more accurately learn the relationships between words and the position in the map incrementally by using method (A).

\section{Conclusion}
\label{sec:conclusion}
This paper discussed online learning methods of spatial concepts and an environmental map by a mobile robot.
The proposed method integrated the spatial concept acquisition into SLAM by an RBPF-based approach.
In the experiments, we conducted online learning in a novel environment by the robot without a pre-existing lexicon and map.
The experimental results demonstrated that SpCoSLAM enhances the performance of place recognition using a speech signal in online learning methods.
SpCoSLAM improved over-segmentation problem in lexical acquisition by updating the language model sequentially.
We consider that incorporating forgetting~\cite{araki2012online} and rejuvenation~\cite{canini2009online} into SpCoSLAM could further improve estimation accuracy.

One of the advantages of online learning is that it can deal with changes in the environment and place names.
Moreover, we consider that the spatial concepts the robot mistakenly learns can also be corrected sequentially.
In this way, it will be possible to acquire spatial concepts that flexibly respond to changes in the environment, which could not be done so far.
We expect this work to contribute greatly to the realization of long-term spatial language interaction between people and robots.
In the future, we would like to perform continuous online learning of spatial concepts in a long-term dynamic environment and incremental transfer learning of spatial concepts to other novel environments. 


\bibliographystyle{IEEEtran}   
\bibliography{ForJabReff_conf} 

\end{document}